\documentclass[runningheads]{llncs}
\usepackage{graphicx}

\usepackage{tikz}
\usepackage{comment}
\usepackage{amsmath,amssymb} 
\usepackage{color}

\usepackage[accsupp]{axessibility}  

\newcommand{\jinyu}[1]{\textcolor{red}{#1}}

\begin{document}
\pagestyle{headings}
\mainmatter
\def\ECCVSubNumber{22}  

\title{Learning Dual-Fused Modality-Aware Representations for RGBD Tracking} 

\titlerunning{Learning Dual-Fused Modality-Aware Representations for RGBD Tracking}
%

\author{
Shang Gao\inst{1} \and
Jinyu Yang\inst{1,2} \and
Zhe Li\inst{1} \and
Feng Zheng\inst{1}\thanks{Corresponding author.} \and
Ale\v{s} Leonardis\inst{2} \and
Jingkuan Song\inst{3}
}
\authorrunning{S. Gao et al.}
\institute{Department of Computer Science and Engineering,\\ Southern University of Science and Technology, Shenzhen, China \and
University of Birmingham, Birmingham, United Kingdom \and
University of Electronic Science and Technology of China , Chengdu, China
}

\maketitle

\begin{abstract}
With the development of depth sensors in recent years, RGBD object tracking has received significant attention. 
Compared with the traditional RGB object tracking, the addition of the depth modality can effectively solve the target and background interference. 
However, some existing RGBD trackers use the two modalities separately and thus some particularly useful shared information between them is ignored.
On the other hand, some methods attempt to fuse the two modalities by treating them equally, resulting in the missing of modality-specific features.
To tackle these limitations, 
we propose a novel Dual-fused Modality-aware Tracker (termed DMTracker) which aims to learn informative and discriminative representations of the target objects for robust RGBD tracking.
The first fusion module focuses on extracting the shared information between modalities based on cross-modal attention.
The second aims at integrating the RGB-specific and depth-specific information to enhance the fused features.
By fusing both the modality-shared and modality-specific information in a modality-aware scheme, our DMTracker can learn discriminative representations in complex tracking scenes.
Experiments show that our proposed tracker achieves very promising results on challenging RGBD benchmarks. 

\end{abstract}

\section{Introduction}

Object tracking is to localize an arbitrary object in a video sequence, given only the object description in the first frame.
It can be applied in lots of applications in video surveillance, autonomous driving \cite{luo2018fast,zheng2021box,qi2020p2b}, and robotics \cite{machida2012human}.
Recent years witness the development of RGBD (RGB+Depth) object tracking thanks to the affordable and accurate depth cameras.
RGBD tracking aims to track the objects more robustly and accurately with the help of depth information, even in color-failed scenarios, \textit{e.g.}, target occlusion and dark scenes.
\begin{figure}[t] 
\centering
\includegraphics[width=\linewidth]{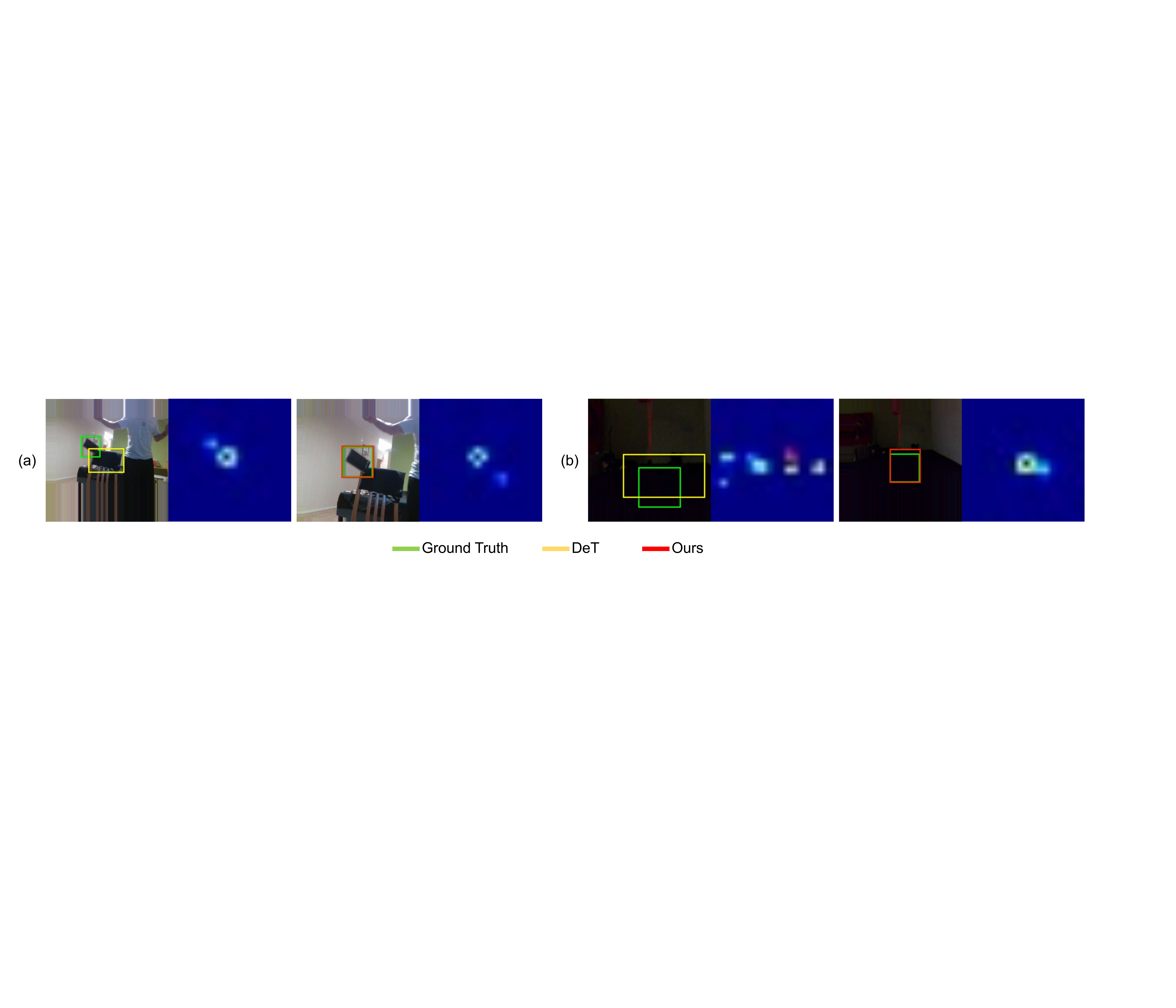}
\caption{Visual comparison of our proposed DMTracker and the state-of-the-art DeT \cite{det}. Our tracker is more robust and accurate on handling the tracking challenges, \textit{e.g.}, (a) background clutter, (b) dark scenes.}
\label{visual}
\end{figure}
Compared to conventional RGB-based tracking, the major difficulty of RGBD tracking is the modality discrepancy resulting from intrinsically distinct information obtained from different modalities.
For accurate and robust tracking, existing RGBD trackers have many attempts on either RGB and depth fusion \cite{ca3dms}, or RGB and depth feature extraction \cite{ptb}.
For example, early RGBD trackers fuse RGB and depth HoG features by using different weights \cite{stc}. 
Camplani \textit{et al.} \cite{dskcf} applied the Kernel Correlation Filter (KCF) in RGB and depth maps, respectively. 
With the boosting of deep learning, researchers combine depth information with pre-trained RGB trackers to improve tracking performance.
DAL \cite{dal} utilizes depth information to enhance RGB features via depth-aware convolutions.
Very recently, Yan \textit{et al.} proposed DeT \cite{det}, which utilized the depth colormaps for feature extraction and fused color and depth features via a simple operation like element-wise maximum.

However, on the one hand, a simple modality fusion strategy generally treats the two modalities equally. 
Thus, some modality-specific features which are strongly complementary cannot be discovered. 
For example, such fusing operations have side effects on the color modality.
As RGB frames are informative, preserving the color and texture information is important for tracking.
On the other hand, separately using RGB and depth information may lose the shared information between different modalities.
Thus, with simply-fused features only, or using color and depth features separately, the upper bound of the discrimination ability of the feature representation is limited. 
Up to now, few methods explicitly exploit both intra- and inter-modality characteristics for RGBD tracking.
Such deficiencies impede the tracking performance, and thus designing a high-performance RGBD tracker is still challenging.

Existing works in multi-modal learning demonstrate that models can benefit from exploring both the shared information and modality-specific characteristics \cite{2017Sharable}.
Inspired by this, in this paper, we aim to tackle the above deficiencies by proposing \textbf{D}ual-fused \textbf{M}odality-aware \textbf{Tracker} for RGBD object tracking, namely \textbf{DMTracker}.
The proposed framework consists of dual fusions, which apply the modality similarity and discrepancy for a robust feature representation.
The first fusion module is the Cross-Modal Integration Module (CMIM), which is based on cross-attention.
The second fusion module is Specificity Preserving Module (SPM).
Specifically, CMIM explores the affinities between RGB and depth features to obtain shared information between different modalities.
The obtained modality-shared features represent the correlation between RGB and depth channels through cross-modal attention.
SPM compensate the modality-shared features by fusing the weighted modality-specific features and the shared ones.
Thus, the second fusion tries to make up the missing specific information from the individual modalities.
In DMTracker, intra-modality and inter-modality characteristics are both preserved through modality-aware fusions for similarity matching between tracking templates and search regions.
This scheme can compensate for the lack of modality-specific information and enhance the modality-shared features, which finally improves the overall representation ability.
Experiments on popular RGBD tracking benchmarks demonstrate the effectiveness of the proposed method.



Our contributions are three-fold:
\begin{itemize}
    \item We propose a new tracking method dedicated to RGBD object tracking. It is capable of end-to-end offline training and real-time applications.
    \item We develop two novel modality-aware fusion modules that incorporate both modality-shared and modality-specific information for robust tracking.
    \item The proposed DMTracker achieves state-of-the-art performance on existing challenging RGBD tracking benchmarks with running at real-time speed.
\end{itemize}

\section{Related Work}
\textbf{RGBD Object Tracking.}
Early RGBD trackers devote to applying RGBD fusion on handcrafted features between color and depth channels.
PTB \cite{ptb} presents a hybrid RGBD tracker composed of HOG features, optical flow, and 3D point clouds.
An \textit{et al.} \cite{dls} extends KCF by adding the depth channel.
Another idea is to utilize the depth information on specific scenarios, \textit{e.g.}, occlusion handling and target re-detection \cite{ca3dms,otr}, with hand-crafted depth features.
In late fused trackers, the decisions from different channels are combined by using weighted summation \cite{stc}, or multi-step localization \cite{mcbt}. 
Recently, deep learning-based RGBD trackers appear \cite{dal,tsdm,det}.
Among them, DAL \cite{dal} reformulates a deep discriminative correlation filter (DCF) to embed the depth information into deep features.
DeT \cite{det} transfers the depth maps to depth colormaps and fuses the color and depth features via a mixing operation, \textit{e.g.}, maximum.
However, these methods ignore the modality discrepancy, which decreases the tracking accuracy.
Compared with existing methods, our DMTracker eliminates the bias that heterogeneous features learned from different modalities, and exploits the correlation between them for robust tracking.

\noindent
\textbf{Multi-modal Learning.}
Multi-modal learning has attracted more and more attention since a large amount of data can be collected from various sources or sensors.
As events or actions can be described by information from multiple modalities, multi-modal learning aims to understand the correlation between different modalities.
Early methods directly concatenate the features from the multiple modalities into a feature vector.
However, simple concatenations often fail to exploit the complex correlations among different modalities.
Therefore, several multi-modal learning methods have been developed to explicitly fuse the complementary information from different modalities to improve model performance \cite{kim2020modality,pan2020spatio}.
Hu \textit{et al.} \cite{2017Sharable} present a shareable and individual multi-view learning strategy to explore more properties of multi-modal data. Besides, Zhou \textit{et al.} \cite{spnet} propose SPNet which improves saliency detection performance by exploring both the shared information and modality-specific properties.
Chen \textit{et al.} \cite{chen2021end} propose a multi-modal framework with an inter-modal module that learns cross-modal features and an intra-modal module to self-learn feature representations across videos.

\noindent
\textbf{Transformer in Tracking.}
The transformer is originally proposed for natural language processing.
With attention-based encoders and decoders, transformers have shown their great potential for vision tasks.
Transformers have been applied to generic object tracking with considerable success \cite{transt,stark,cao2022tctrack,yu2021high,lin2021swintrack,wang2021transformer}.
Existing transformers for tracking focus on the correlation between target templates and search regions.
Typically, the transformer decoder fuses feature from the template and search region by cross-attention layers to obtain discriminative features.
For example, TransT \cite{transt} is proposed to use cross-attention instead of the traditional correlation operation. 
Later, STARK is proposed based on DETR \cite{detr}, utilizing self-attention to learn the relationship between different inputs.
Existing works demonstrate the effectiveness of the transformer on track, while they mostly focus on the RGB tracking area. 
To the best of our knowledge, there are no transformer architectures focusing on cross-modal RGBD tracking.
Our proposed method utilizes an attention mechanism between color and depth modalities and shows promising performance.

\begin{figure*}[t] 
\centering
\includegraphics[width=\textwidth]{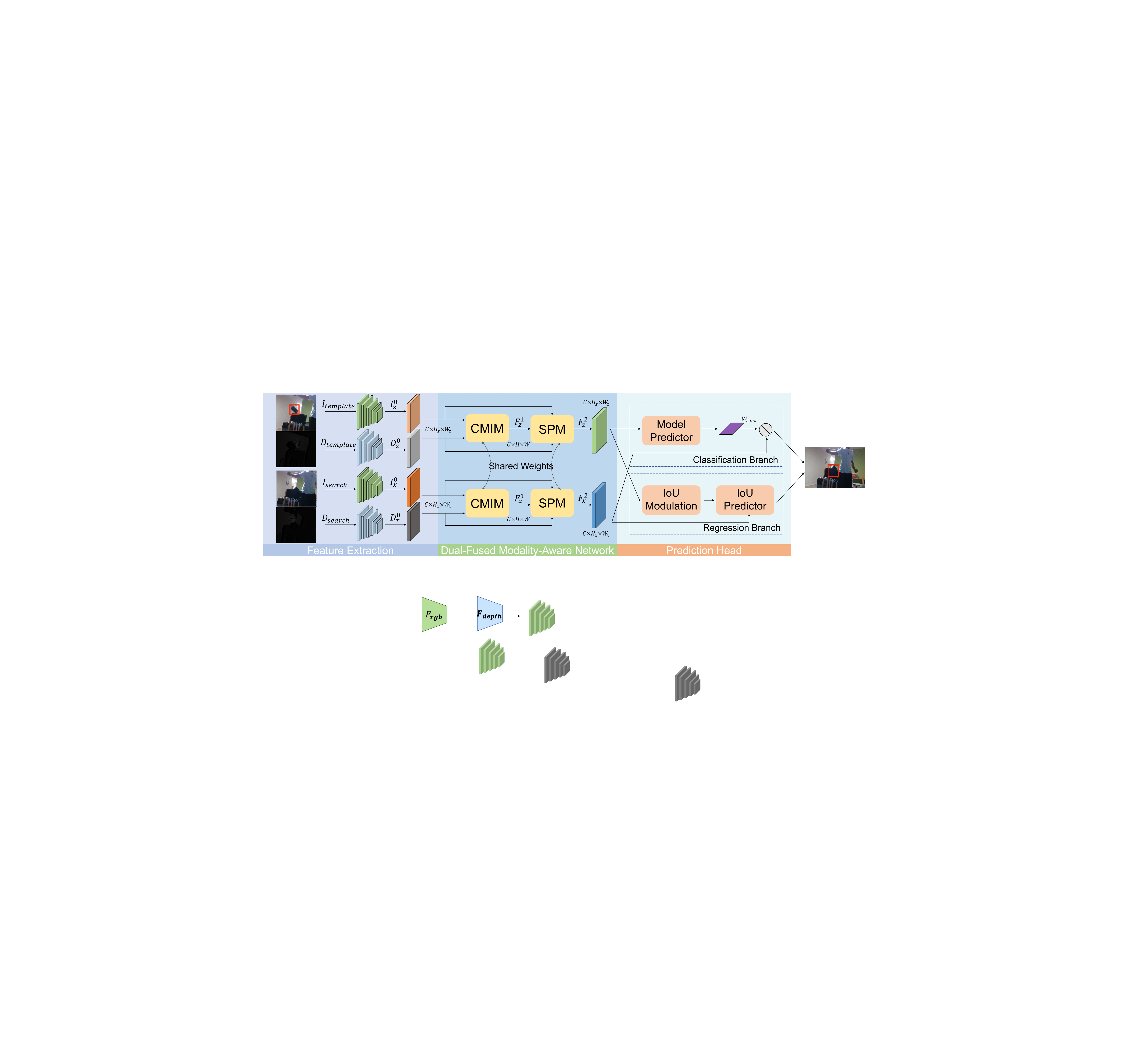}
\caption{Overview of the DMTracker pipeline. Our DMTracker consists of three main components: feature extraction, dual-fused modality-aware network, and prediction head.
The dual-fused modality-aware network consists of two novel modules, CMIM (cross-modal integration module) and SPM (specificity preserving module), which are introduced in Section \ref{fusion1} and \ref{fusion2}. }
\label{structure}
\end{figure*}
\section{Methodology}
In this section, we present the cross-modal correlation tracker, \textit{i.e.} DMTracker. 
We first describe our DMTracker pipeline in Section~\ref{overview}.
Next, we present the key modules in the dual-fused modality-aware network in Section~\ref{fusion1} and \ref{fusion2}, in which the two main modules will be introduced.
Training loss is present in Section~\ref{training}.
Finally, we detail our training and inference procedures in Section~\ref{imple}.


\subsection{Overview}\label{overview}
As shown in Figure \ref{structure}, the proposed DMTracker contains three main components: feature extractor, dual-fused modality-aware network, and prediction head.
The feature extractor separately extracts the features from the template and the search region of both RGB and depth modalities. 
Then, the features from different modalities are enhanced by the proposed dual-fused modality-aware network to be a robust representation.
Finally, the prediction head performs the binary classification and bounding box regression on the fused features to predict the bounding boxes.


\noindent
\textbf{Feature extractor.} 
Firstly, the feature extractor takes template patches from RGB and depth frames $Z_{rgb},~Z_{depth} \in \mathbb{R}^{3\times H_{z_{0}}\times W_{z_{0}}}$ and corresponding search region patches $X_{rgb},~X_{depth} \in \mathbb{R}^{3\times H_{x_{0}}\times W_{x_{0}}}$.
With the ResNet-50 backbone, the feature maps of the template and the search patch from the two modalities, \textit{i.e.}, $I_{z},~D_{z} \in \mathbb{R}^{C\times H_{z}\times W_{z}}$ and $I_{x}, D_{x} \in \mathbb{R}^{C\times H_{x}\times W_{x}}$, can be obtained. 
$H_{z}= \frac{H_{z_{0}}}{n}$, $ W_{z} = \frac{W_{z_{0}}}{n}$, $H_{x} = \frac{H_{x_{0}}}{n}$, $W_{x} = \frac{W_{x_{0}}}{n}$, $n$ is the downsampling factor and $C$ is the dimension of the feature maps.

\noindent
\textbf{Dual-fused modality-aware network.}
Then, the dual-fused modality-aware network takes the feature maps of the two modalities of RGB and depth under the same branch as input, \textit{i.e.}, $I_{x}, D_{x}, I_{z}$, and $D_{z}$.
The whole network consists of two fusion modules, \textit{i.e.}, the Cross-Modal Integration Module (CMIM) and Specificity Preserving Module (SPM).
Features are firstly fed into the preliminary fusion in CMIM to obtain the initial fusion features through the modality-aware cross-modal attention.
And then the features, and both the RGB and depth features, are fed into the SPM to get the final fused features with preserving the modality-aware specificity.
The details of the two modules in the network are introduced in Section \ref{fusion1} and \ref{fusion2}.

\noindent
\textbf{Prediction head.}
After getting the final modality-aware feature representations of templates and search regions, a prediction head \cite{dimp} is used to obtain the bounding box estimation.
For the classification head, the fused template feature maps are input to the model predictor that fully exploits the information from the target and background. 
The model predictor outputs the weights of the convolution layer, which then performs target classification for the search branch.
For the localization head, using the template features and the initial target bounding box, IoU Modulation component computes the modulation vectors carrying the target-specific appearance information.
Finally, bounding box estimation is performed by maximizing the IoU prediction based on the modulation vectors and proposal bounding boxed in the search region.

\subsection{Cross-Modal Integration Module (CMIM)}\label{fusion1}
\begin{figure}[t]
  \centering
  \begin{minipage}{0.36\linewidth}
\centering
\includegraphics[width=\linewidth]{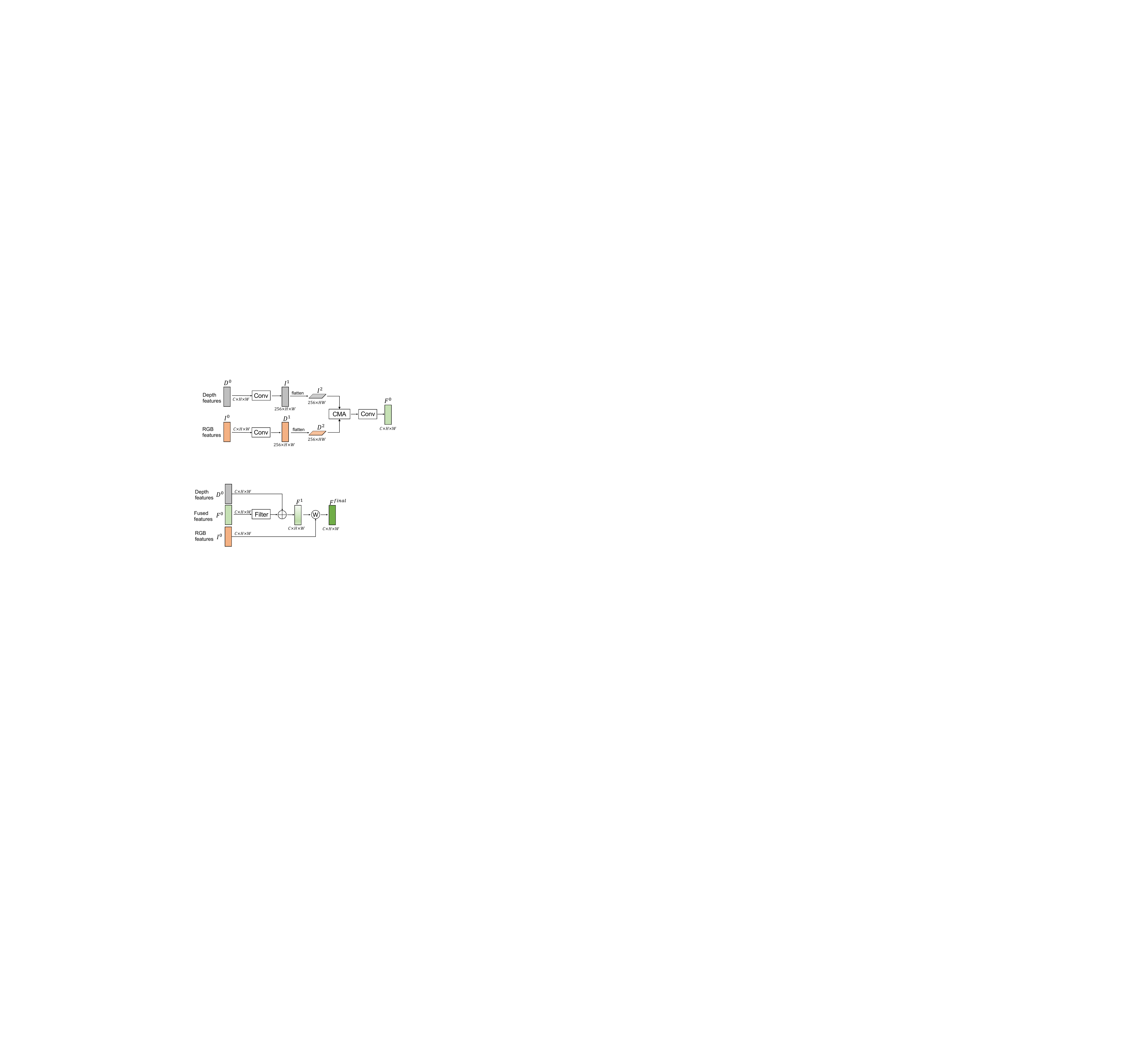}
\caption{Cross-Modal Integration Module (CMIM).}
\label{cross-attention}
  \end{minipage}
~  
  \begin{minipage}{0.36\linewidth}
\centering
\includegraphics[width=0.8\linewidth]{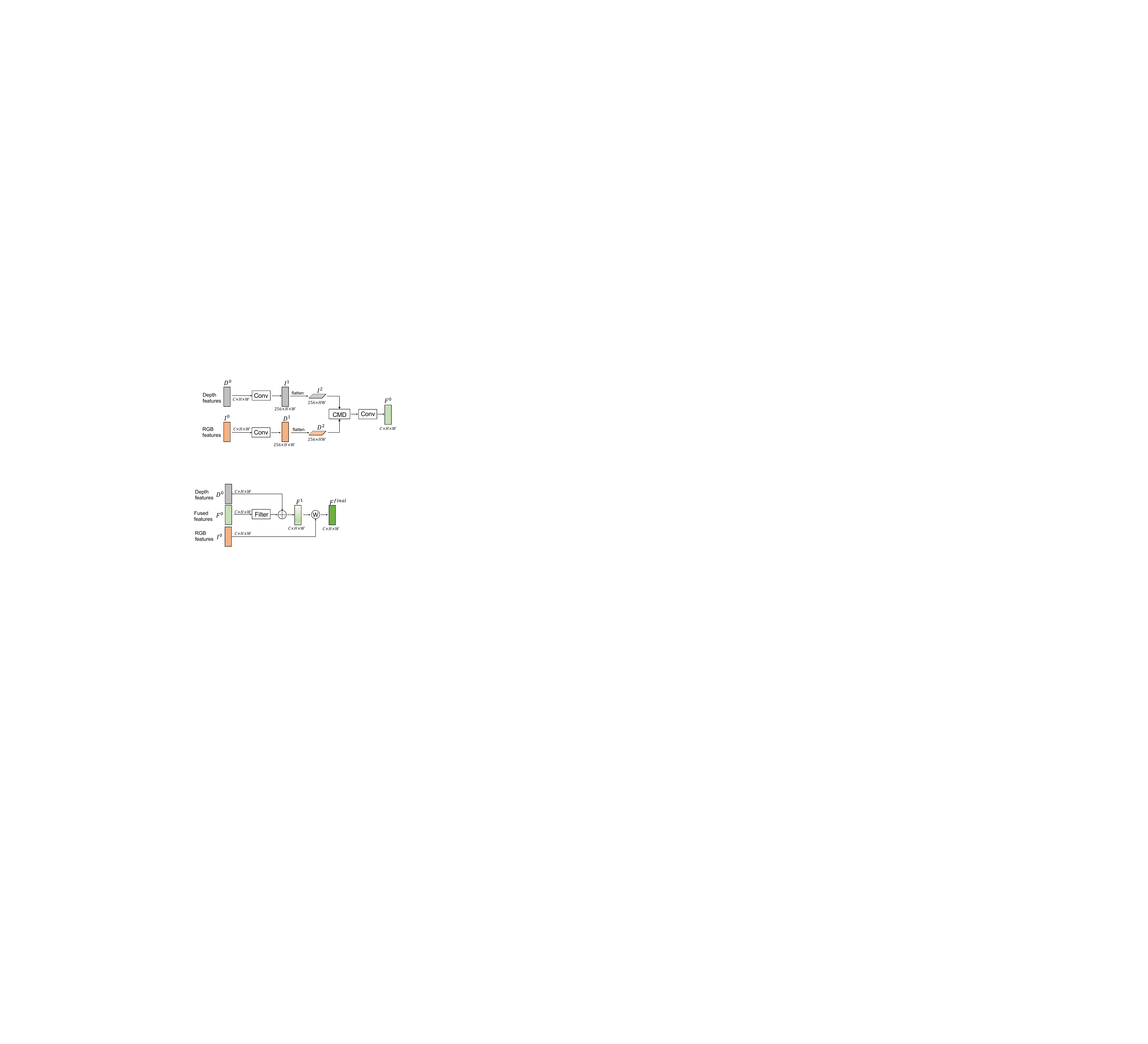}
\caption{Specificity Preserving Module (SPM).}
\label{spm}
  \end{minipage}
\centering
  \begin{minipage}{0.18\linewidth}
\includegraphics[width=\linewidth]{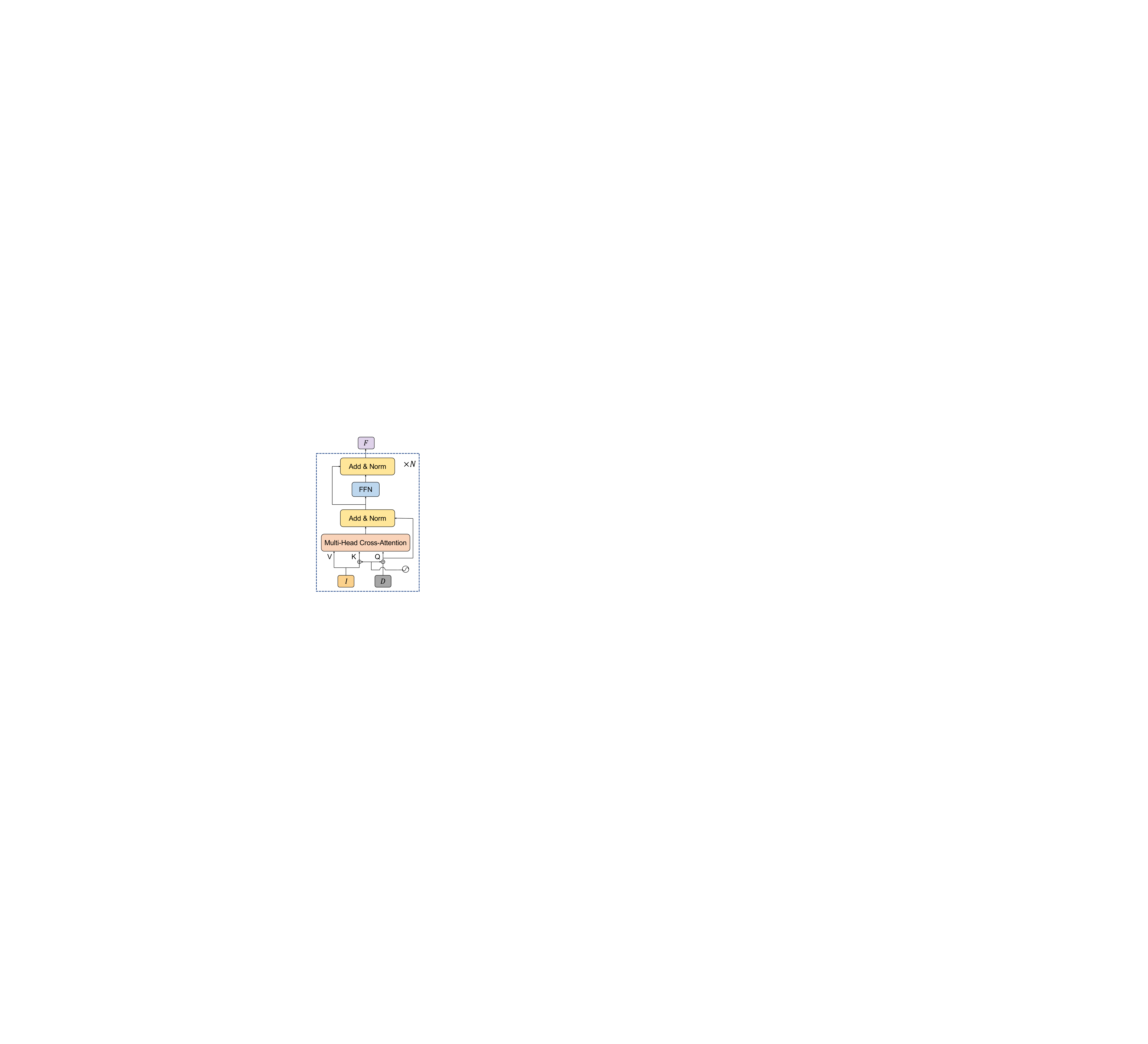}
\caption{Cross-Modal Attention (CMA).}
\label{cma}
  \end{minipage}
\end{figure}

The cross-modal integration module is built as shown in Figure \ref{cross-attention}.
In this module, we fuse the cross-modal features from the RGB and depth modalities to learn their shared representation.

\noindent
\textbf{Attention.} 
In the attention mechanism, given queries $Q$, keys $K$ and values $V$, the attention function is defined by the scale dot-product attention: 
\begin{equation}
    Attention(Q,K,V) = Softmax(\frac{QK^\top}{\sqrt{d_{k}}})V,
\end{equation}
where $d_{k}$ is the dimension of the key $K$.

It can also be extended to multi-head attention (MHA) as: 
\begin{equation}
\begin{split}
    MHA(Q,K,V) = Concat(Head_{1},Head_{h})W^{O}, \\
    Head_{i} = Attention(QW^{Q}_{i}, KW^{K}_{i}, VW^{V}_{i}),
\end{split}
\end{equation}
where the projections matrices are $W^{Q}_{i}, W^{K}_{i} \in \mathbb{R}^{d_{model} \times d_{k}}$, $W^{V}_{i}\in\mathbb{R}^{d_{model} \times d_{v}}$ and $W^{O} \in \mathbb{R}^{hd_{v}\times d_{model}}$.
By projecting the inputs to different spaces through different linear transformations, the model can learn various data distributions and pay attention to varying levels of information, for more details \cite{attention}.
In this work, we employ $h$ = 8, $d_{model}$ = 256 and $d_{k} = d_{v} = d_{m} = 32$ as default values.

\noindent
\textbf{Cross-Modal Attention (CMA) Block.}
To model the global information of the dual modalities and minimize the computation of our real-time tracker, 
we use the lightweight attention architecture named cross-modal attention (CMA) block as shown in Figure \ref{cma}. We simplify the original transformer, remove the self-attention part, but retain the cross-attention, and feed-forward network because we pay more attention to the interactive fusion between the dual modalities, and it can significantly reduce the amount of computation and improve the speed of the tracker. A feed-forward network (FFN) is a fully connected network that consists of two linear functions with a Relu activate function to enhance the fitting ability of the attention. The attention mechanism can not distinguish the position information of the input feature sequence. 
Thus, we introduce the spatial positional encoding attracting to input following the setting \cite{detr,transt}. 
Cross-Modal Attention Block can enhance depth features by finding the most related visual clues from RGB features, and the CMA can be formulated as equation($3$).
\begin{equation}
\begin{split}
    f_{t} &= Norm(D+MHA(D, I, I)),\\
    F &= Norm(f_{t}+FFN(f_{t})),
\end{split}
\end{equation}
where $F, D \in \mathbb{R}^{C \times H \times W}$.

\noindent
\textbf{Cross-Modal Integration.}
The cross-modal integration module is designed based on the CMA block. 
The basic idea of our proposed CMIM is to use the attention mechanism to globally model the correlation information between the dual modalities learning the shared representation.
Formally, we use $I^{(0)},D^{(0)}$ to denote the initial RGB feature and depth feature obtained from the backbone network. First, a 1x1 convolution reduces the channel dimension obtaining two lower dimension feature $I^{(1)}$ and $D^{(1)}$:
\begin{equation}
    I^{(1)}, D^{(1)} = \mathcal{F}(I^{(0)},D^{(0)}),
\end{equation}
where $I^{(0)},D^{(0)}\in \mathbb{R}^{C \times H \times W},I^{(1)},D^{(1)}\in \mathbb{R}^{C_{i} \times H \times W}$. 
We employ $C_{i} = 256$ in our implementation. 
Since the cross-modal attention (CMA) takes a set of feature vectors as input, we flatten $I^{(1)}$ and $D^{(1)}$ in spatial dimension, obtaining $I^{(2)}\in \mathbb{R}^{C_{i}\times HW }$ and $D^{(2)}\in \mathbb{R}^{C_{i}\times HW }$. 
In our scenario, we have dual modality features which can be used to attend to each other using the Dual-fused Modality-aware Network \textit{i.e.}, using RGB feature to enhance depth feature, or using depth feature to enhance RGB feature.
In our method, we adopt the former: $D$ as query, $I$ as key and value, as this essentially puts more emphasis on the RGB feature, if we want to make depth feature to integrate more color or texture features from the RGB feature. 
After cross-modal attention (CMA) block and a 1x1 convolution, the initial fused RGBD feature $F^{(0)}$ can be obtained as follows:
\begin{equation}
\begin{split}
    F^{(0)} &= \mathcal{F}(CMA(I^{(2)},D^{(2)})), \quad F^{(0)} \in \mathbb{R}^{C\times H \times W}.
\end{split}
\end{equation}

\subsection{Specificity Preserving Module (SPM)}\label{fusion2}
The specificity preserving module is built as shown in Figure \ref{spm}.
As we have the shared representations from the fused modality, and the features from the modality-specific feature extractors, we propose SPM that make full use of modality-specific information to further enhance the fusion features. Due to the considerable gap between the depth space information from the depth feature and the visual information from the RGB feature, the initial fused features obtained by the CMIM not only contain the correlation information between the two modalities but are also mixed with a lot of irrelevant interference features. Thus, the initial fused feature first passes through an information filter to weaken the chaotic interference information and enhance the shared characteristics of the dual modalities. Then the specificity of each modality is attached to the filtered features, and the final fused features can be obtained which contain the modality-shared and modality-specific cues. The detailed process is described as follows.

We design a learnable vector $\mathcal{V}$ to filter the interference and enhance the shared cues of the initial fused features. $\odot$ means element-wise multiplication.
It is then updated with depth-specific features through residual connections:
\begin{equation}
    F^{(1)} = D^{(0)}+\mathcal{V} \odot F^{(0)}, \mathcal{V} \in \mathbb{R}^{C}, F^{(0)} \in \mathbb{R}^{C\times H \times W}.
\end{equation}

Then the obtained features $F^{(1)}$ and initial RGB-specific features $I^{(0)}$ are fused by learnable weights to get the final RGBD feature($7$).
\begin{equation}
    F^{(final)} = \alpha I^{(0)}+\beta F^{(1)} , \quad F^{(final)} \in\mathbb{R}^{C\times H \times W}.
\end{equation}


\subsection{Training Loss}\label{training}
We train our network with the following function,
\begin{equation}
    \mathcal{L}_{total} = \lambda \mathcal{L}_{cls} + \mathcal{L}_{bbox}.
\end{equation}
Classification loss $\mathcal{L}_{cls}$ provides the network the ability to robustly discriminate the target object and background distractors, which is defined as \cite{dimp}.

\begin{equation}
l(s,z)=\left\{
\begin{array}{rcl}
s-z & & {z > T}\\
max(0,s) & & {z \leq T}
\end{array} \right.
\end{equation}
\begin{equation}
\mathcal{L}_{cls} = \frac{1}{N_{iter}}\sum_{i=0}^{N_{iter}}\sum_{(x,c)\in M_{test}} \Vert l(x \ast f, z_{c}) \Vert^2.
\end{equation}
Here, according to the label confidence $z$, the threshold $T$ defines the target and background regions. $N_{iter}$ refers to the number of iterations. $z_{c}$ is the target center label that is set from a Gaussian function. We only penalize positive confidence values for the background if $z \leq T$. If $z > T$, we take the difference between them. Here, $f$ is the wights obtained by the model predictor network. 

For bounding box estimation, following \cite{atom}, it extends the training procedure to image sets by computing the modulation vector on the first frame in $M_{train}$ and sampling candidate boxes from images in $M_{test}$. The bounding box loss $\mathcal{L}_{bbox}$ is computed as the mean squared error between the predicted BBox defined by $B$ and the groundtruth BBox defined by $\bar B$ in $M_{test}$,

\begin{equation}
    \mathcal{L}_{bbox} = \mathop{MSE} \limits_{i\in M_{test}}(B_{i}, \bar B_{i}).
\end{equation}

\subsection{Implementation Details}\label{imple}
\noindent
\textbf{Training.}
Our tracker is trained in an end-to-end fashion.
1) Template and Search Area. Different from DeT \cite{det}, we use RGB images and raw depth images as inputs. To adapt the depth image to the pre-trained feature extraction model, we copy the raw depth to stack three channels. To incorporate background information, the template and search region is obtained by adding random translation and cropping a square patch centered on the target, which is $ 5^{2}$ of the target area. Then these regions are resized to $288 \times 288$ pixels. Horizontal flip and brightness jittering are used for data augmentations.
2) Initialization.
The backbone ResNet50 is initialized by the pre-trained wights on ImageNet. The weights of IoU modulation and IoU predictor in the regression head are initialized using \cite{atom}. The model predictor in the classification head is initialized using \cite{dimp}. $N_{iter}=5, T=0.05.$ We use $\mathcal{V}=0.01$ to initialize the information filter. The learnable weights between the two modalities:$\alpha$ and $\beta$ are initialized to 0.5. The network is optimized using the AdamW optimizer with a learning rate decay of 0.2 every 15th epoch and weight decay $1e-4$. The target classification loss weight is set to $\lambda$ = $1e-2$. 

\noindent
\textbf{Inference.}
Different from the training phase, we track the target in a series of subsequent frames giving the target position in the first frame. Once the annotated first frame is given, we use the data enhancement strategy to build an initial set of 15 samples to refine the model predictor. With the subsequent frames with high confidence, the model predictor can continuously refine the classification head and enhance the ability to distinguish the target and background. Bounding box estimation is performed as in the training phase.

\subsection{Visualization}\label{visualization}
\begin{figure}[t] 
\centering
\includegraphics[width=\linewidth]{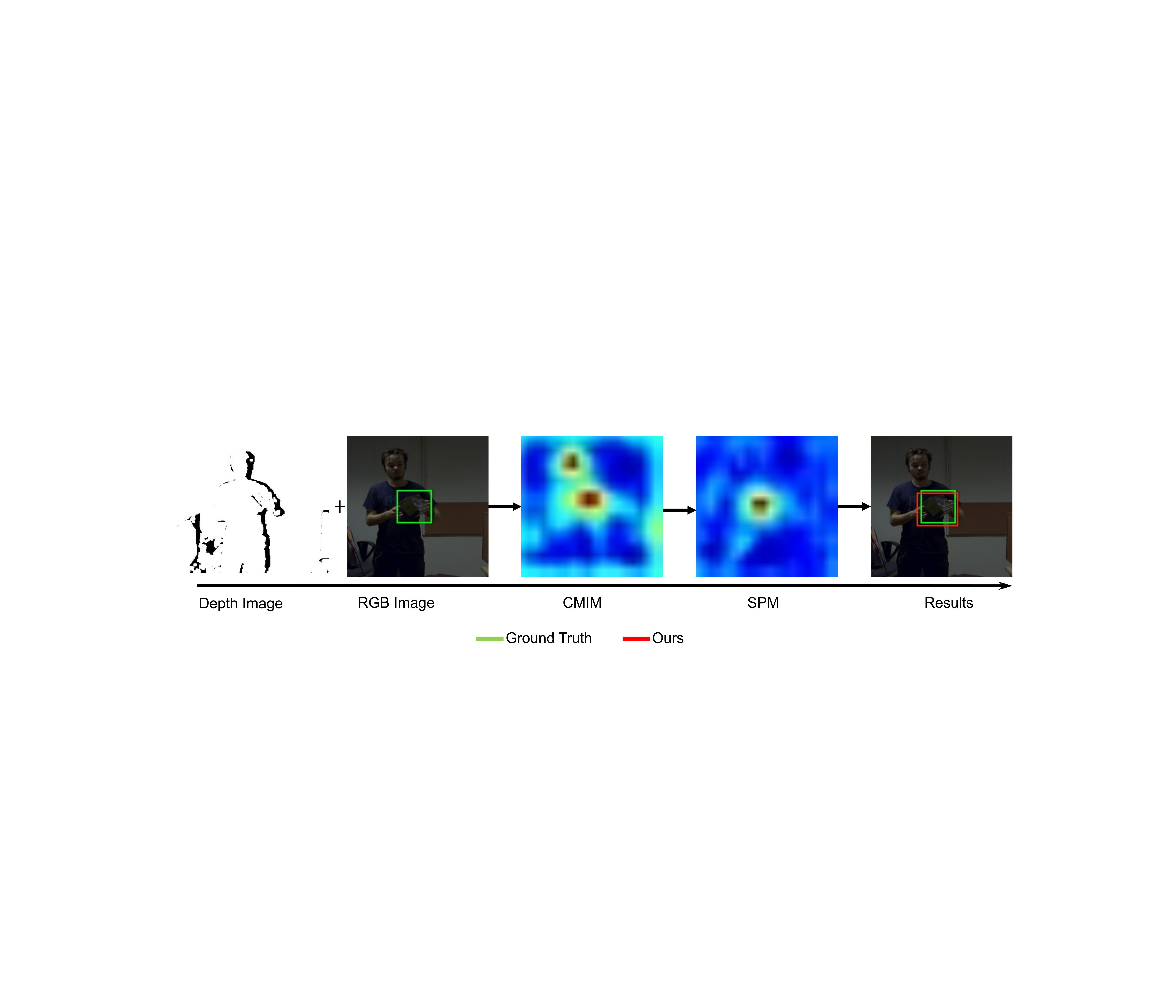}
\caption{Visualization of the attention maps for a representative frame. 
From left to right, they are input depth and RGB frames, feature maps in the search region after CMIM and SPM, and the output prediction, respectively.}
\label{attention map}
\end{figure}

To explore how the dual-fused modality-aware network works in our framework, we visualize the attention maps of the fusion modules in a representative frame, as shown in Figure \ref{attention map}.
We visualize the attention maps of the modality shared and specific features from the output of CMIM and SPM, respectively.
As shown, we can obtain a very informative feature map after the proposed CMIM.
Based on cross-modal attention, the modality-shared feature is generated to make the focus on candidates, \textit{i.e.}, the box and the face.
Then the SPM is to use the modality-specific information to suppress unreliable candidates.
The combination of two kinds of features in SPM produces high-quality representations. 
This proves that the shared and specific features can be well fused by our CMIM and SPM, leading to more discriminative feature representations for tracking.
\section{Experiments}
\subsection{Experiment Settings}




\subsubsection{Datasets.}
To validate the effectiveness of the proposed tracker, we evaluate it on three public RGBD tracking datasets, including STC \cite{stc}, CDTB \cite{cdtb} and DepthTrack \cite{det}.
Among them, STC is a compact short-term benchmark containing 36 sequences, while CDTB and DepthTrack are long-term ones containing 80 sequences and 50 sequences, respectively.
Furthermore, we carry out an attribute-based evaluation to study the performance under different challenges.

\subsubsection{Evaluation metrics.}
For STC, a representative short-term evaluation protocol, success rate, is used.
Given the tracked bounding box $b_t$ and the groundtruth bounding box $b_a$, the overlap score is defined as $S = \frac{b_t \cap b_a}{b_t \cup b_a}$, which is indeed the intersection and union of two regions.
With a series of thresholds from 0 to 1, the area-under-curve (AUC) is calculated as the success rate.

For CDTB and DepthTrack, long-term evaluation, \textit{i.e.}, tracking precision and recall are used \cite{yang2022rgbd}.
They can be obtained by:
\begin{equation} \label{eq:pr_re_f_frames_based}
Pr(\tau _{\theta }) = \frac{1}{N_{p}} \sum_{A_{t}(\tau _{\theta })\neq \varnothing } \Omega (A_{t}(\tau _{\theta }),G_{t}), ~~ 
Re(\tau _{\theta }) = \frac{1}{N_{g}} \sum_{G_{t}\neq \varnothing }  \Omega (A_{t}(\tau _{\theta }),G_{t}),
\end{equation}
where $Pr(\tau_{\theta})$ and $Re(\tau_{\theta})$ denote the precision (\textit{Pr}) and the recall (\textit{Re}) over all frames, respectively. F-score then can be obtained by $F(\tau _{\theta })  =\frac{2Re(\tau _{\theta })Pr(\tau _{\theta })}{Re(\tau _{\theta })+Pr(\tau _{\theta })}$.
$N_{p}$ denotes the number of frames in which the target is predicted visible, and $N_{g}$ denotes the number of frames in which the target is indeed visible.

\subsubsection{Compared trackers.}
We compare the proposed tracker with the following benchmarking RGBD trackers.
\begin{itemize}
    \item Traditional RGBD trackers: PT \cite{ptb}, OAPF \cite{oapf}, DS-KCF \cite{dskcf}, DS-KCF-shape \cite{dsshape}, CA3DMS \cite{ca3dms}, CSR\_RGBD++ \cite{csr}, STC \cite{stc} and OTR \cite{otr};
    \item Deep RGBD trackers: DAL \cite{dal}, TSDM \cite{tsdm}, and DeT\footnote{For a fair comparison, we use DeT-DiMP50-Max checkpoint in all experiments.} \cite{det};
    \item State-of-the-art RGB trackers: ATOM \cite{atom}, DiMP \cite{dimp}, PrDiMP \cite{prdimp}, KeepTrack \cite{keeptrack} and TransT \cite{transt}.
\end{itemize}

\begin{table}[t]
\scriptsize
\begin{minipage}{0.48\linewidth}
\caption{Overall performance on DepthTrack test set \cite{det}. The top 3 results are shown in {\color{red}red}, {\color{blue}blue}, and {\color{green}green}.}
\centering
\label{tbl_depthtrack}
\begin{tabular}{l|c|c|c|c}
\hline
 Method  & Type &\multicolumn{1}{c|}{Pr} & \multicolumn{1}{c|}{Re} & \multicolumn{1}{c}{F-score} \\
\hline
DS-KCF~\cite{dskcf} & RGBD &0.075&0.077&0.076 \\
DS-KCF-Shape~\cite{dsshape} & RGBD &0.070 &0.071 &0.071\\
CSR\_RGBD++ \cite{csr} &RGBD & 0.113 &0.115 &0.114  \\
CA3DMS \cite{ca3dms} &RGBD & 0.212 &0.216 &0.214\\
DAL \cite{dal} &RGBD &0.478 &0.390 &0.421 \\
TSDM \cite{tsdm} & RGBD & 0.393 &0.376 &0.384\\
DeT \cite{det} & RGBD & \textcolor{blue}{0.560} & \textcolor{green}{0.506} &\textcolor{blue}{0.532} \\
DiMP50 \cite{dimp} & RGB & 0.387 &0.403 &0.395 \\
KeepTrack \cite{keeptrack} & RGB &\textcolor{green}{0.504} &\textcolor{blue}{0.514} &\textcolor{green}{0.508} \\
TransT \cite{transt} & RGB &0.484 &0.494 &0.489 \\
\hline
\textbf{DMTracker} & RGBD &\jinyu{0.619} &\jinyu{0.597} &\jinyu{0.608} \\
\hline
\end{tabular}
\end{minipage}
~
\begin{minipage}{0.5\linewidth}
\centering
\caption{Overall performance on CDTB dataset \cite{cdtb}. The top 3 results are shown in {\color{red}red}, {\color{blue}blue}, and {\color{green}green}.}
\label{tbl_cdtb}
\begin{tabular}{l|c|c|c|c}
\hline
 Method  & Type &\multicolumn{1}{c|}{Pr} & \multicolumn{1}{c|}{Re} & \multicolumn{1}{c}{F-score} \\
\hline
DS-KCF~\cite{dskcf} &RGBD &0.036 &0.039 &0.038 \\
DS-KCF-Shape~\cite{dsshape} &RGBD &0.042 &0.044 &0.043\\
CA3DMS \cite{ca3dms} &RGBD &0.271 &0.284 &0.259 \\ 
CSR\_RGBD++ \cite{csr} &RGBD & 0.187 &0.201 &0.194 \\
OTR \cite{otr} &RGBD & 0.336 &0.364 &0.312\\
DAL \cite{dal} &RGBD & \textcolor{blue}{0.661} &\textcolor{green}{0.565} &\textcolor{green}{0.592}\\
TSDM \cite{tsdm} & RGBD & 0.578 & 0.541 &0.559\\
DeT \cite{det} & RGBD &\textcolor{green}{0.651}  &\textcolor{blue}{0.633} &\textcolor{blue}{0.642}\\
ATOM \cite{atom} & RGB & 0.541 &0.537 &0.539\\
DiMP50 \cite{dimp} & RGB & 0.546 &0.549 &0.547\\
\hline
\textbf{DMTracker} & RGBD &\jinyu{0.662} &\jinyu{0.658} &\jinyu{0.660} \\
\hline
\end{tabular}
\end{minipage}
\end{table}

\begin{table*}
\scriptsize
\centering
\caption{Overall performance and attribute-based results on STC \cite{stc}. The top 3 results are shown in {\color{red}red}, {\color{blue}blue}, and {\color{green}green}.}\label{tbl_stc}
\begin{tabular}{l |c |c | cccccccccc}
\hline
&  & & \multicolumn{10}{c}{Attributes} \\
Method & Type &Success &\multicolumn{1}{c}{IV} & \multicolumn{1}{c}{DV} & \multicolumn{1}{c}{SV} & \multicolumn{1}{c}{CDV} & \multicolumn{1}{c}{DDV} & \multicolumn{1}{c}{SDC} & \multicolumn{1}{c}{SCC} & \multicolumn{1}{c}{BCC} & \multicolumn{1}{c}{BSC} & \multicolumn{1}{c}{PO} \\
\hline
OAPF \cite{oapf}& RGBD &0.26&0.15&0.21&0.15&0.15&0.18&0.24&0.29&0.18&0.23&0.28\\
DS-KCF \cite{dskcf}& RGBD &0.34&0.26&0.34&0.16&0.07&0.20&0.38&0.39&0.23&0.25&0.29\\
PT \cite{ptb} & RGBD &0.35&0.20&0.32&0.13&0.02&0.17&0.32&0.39&0.27&0.27&0.30\\
DS-KCF-Shape \cite{dsshape} & RGBD &0.39&0.29&0.38&0.21&0.04&0.25&0.38&0.47&0.27&0.31&0.37\\
STC \cite{stc} & RGBD &0.40&0.28&0.36&0.24&0.24&0.36&0.38&0.45&0.32&0.34&0.37\\
CA3DMS \cite{ca3dms} & RGBD &0.43&0.25&0.39&0.29&0.17&0.33&0.41&0.48&0.35&0.39&0.44\\
CSR\_RGBD++ \cite{csr}& RGBD &0.45&0.35&0.43&0.30&0.14&0.39&0.40&0.43&0.38&0.40&0.46\\
OTR \cite{otr} & RGBD &0.49&0.39&0.48&0.31&0.19&0.45&0.44&0.46&0.42&0.42&0.50\\
DAL \cite{dal} & RGBD &\textcolor{blue}{0.62} &\textcolor{blue}{0.52}  &0.60 &\textcolor{red}{0.51} &	\textcolor{blue}{0.62} &	0.46 &	\textcolor{blue}{0.63} &	\textcolor{green}{0.63} &	\textcolor{red}{0.55} &	\textcolor{red}{0.58} &	0.57 \\
TSDM \cite{tsdm} & RGBD &0.57 &\textcolor{red}{0.53}  &0.42 &0.43 &	0.54 &	0.37 &	0.56 &	0.55 &	0.41 &	0.45 &	0.53 \\
DeT \cite{det}  & RGBD &0.60 &0.39  &0.56 &\textcolor{blue}{0.48} &	\textcolor{green}{0.58} &	0.40 &	0.59 &	0.62 &	0.48 &	0.46 &	0.54 \\ 
DiMP \cite{dimp}& RGB &0.60 &0.49 &\textcolor{green}{0.61} &\textcolor{green}{0.47} &0.56 &\textcolor{green}{0.57} &0.60 &\textcolor{blue}{0.64} &0.51 &\textcolor{blue}{0.54} &0.57 \\
PrDiMP \cite{prdimp} & RGB &0.60 &0.47 &\textcolor{blue}{0.62} &	\textcolor{green}{0.47} &	0.51 &	\textcolor{blue}{0.58} &	\textcolor{red}{0.64} &	0.62 &	\textcolor{blue}{0.53} &	\textcolor{blue}{0.54} & \textcolor{blue}{0.58} \\
KeepTrack \cite{keeptrack}& RGB & \textcolor{blue}{0.62} & 0.51 &\textcolor{red}{0.63} &	\textcolor{green}{0.47} &	0.57 &	0.56 &	\textcolor{green}{0.61} &	0.62 &	\textcolor{blue}{0.53} &	\textcolor{green}{0.52} & \textcolor{blue}{0.58}  \\
\hline
\textbf{DMTracker} & RGBD & \jinyu{0.63} & \textcolor{blue}{0.52} &0.59  &0.46 &\jinyu{0.67} &\jinyu{0.63} &	\textcolor{green}{0.61} &	\jinyu{0.65} &	\textcolor{green}{0.52} &	\textcolor{green}{0.52} &\jinyu{0.60} \\
\hline
\end{tabular}
\end{table*}



\begin{figure}[t]
  \centering
  \begin{minipage}{0.48\linewidth}
\centering
\includegraphics[width=\linewidth]{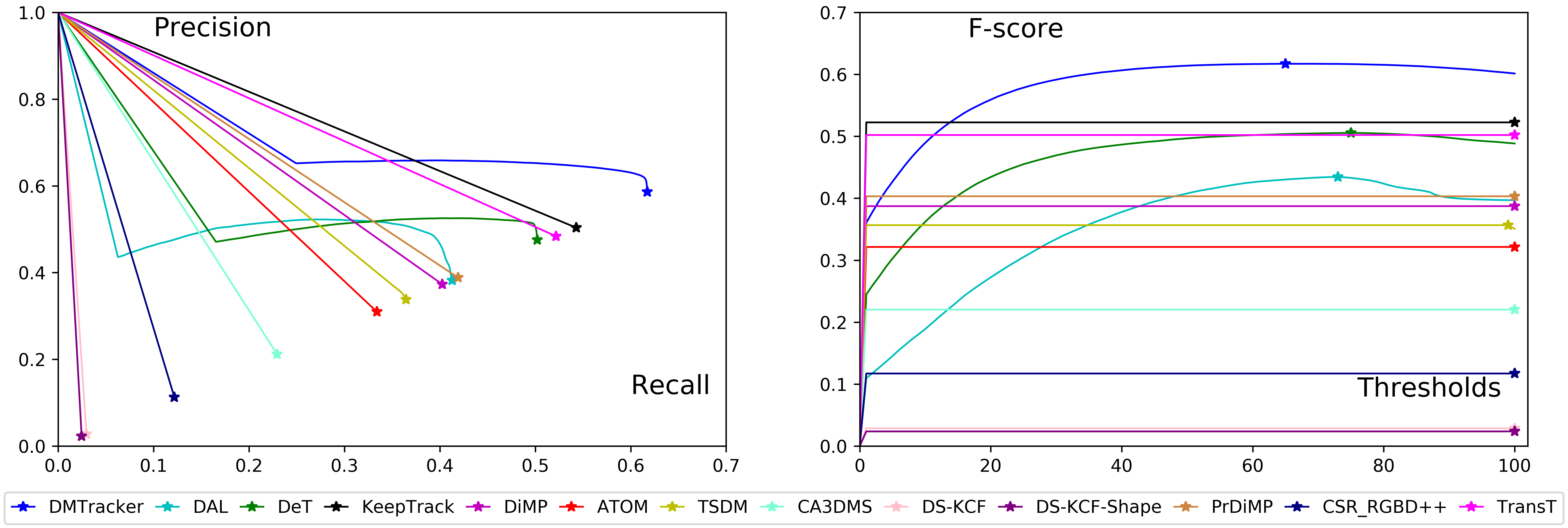}
\caption{Overall tracking performance is presented as tracking Precision-Recall and F-score Plots on DepthTrack
\cite{det}.}
\label{F1 score on depthtrack}
  \end{minipage}
~  
  \begin{minipage}{0.48\linewidth}
\centering
\includegraphics[width=\linewidth]{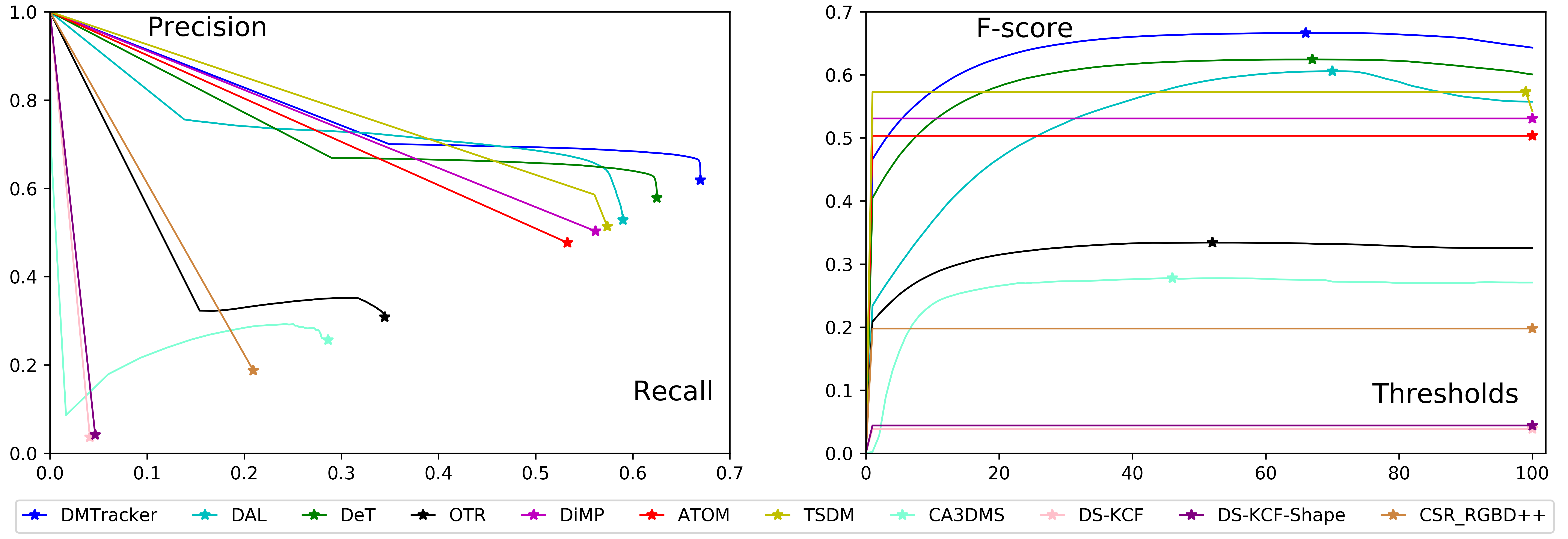}
\caption{Overall tracking performance is presented as tracking Precision-Recall and F-score Plots on CDTB \cite{cdtb}.}
\label{F1 score on cdtb}
  \end{minipage}
\end{figure}

\subsection{Main Results}

\subsubsection{Quantitative results.}
We give the quantitative results of the compared models on DepthTrack, CDTB, and STC in Table \ref{tbl_depthtrack}, \ref{tbl_cdtb}, and \ref{tbl_stc}, respectively.
For DepthTrack and CDTB, corresponding tracking Precision-Recall and F-score Plots are also given in Figure \ref{F1 score on depthtrack} and Figure \ref{F1 score on cdtb}.
As shown in Table \ref{tbl_depthtrack}, our method outperforms all of the compared state-of-the-art methods and obtains the best F-score of 0.608 on the DepthTrack dataset.
On CDTB, we also obtain the top F-score of 0.660.
Compared with the state-of-the-art RGBD tacker DeT, our DMTracker performs 7.6\% and 1.8\% higher on DepthTrack and CDTB, respectively, according to F-score.
DMTracker also outperforms TransT by 11.9\% higher on DepthTrack.
Overall, the proposed DMTracker provide better accuracy than RGB trackers or advanced RGBD trackers on existing RGBD datasets.

\subsubsection{Qualitative results.}
\begin{figure*}[t] 
\centering
\includegraphics[width=\linewidth]{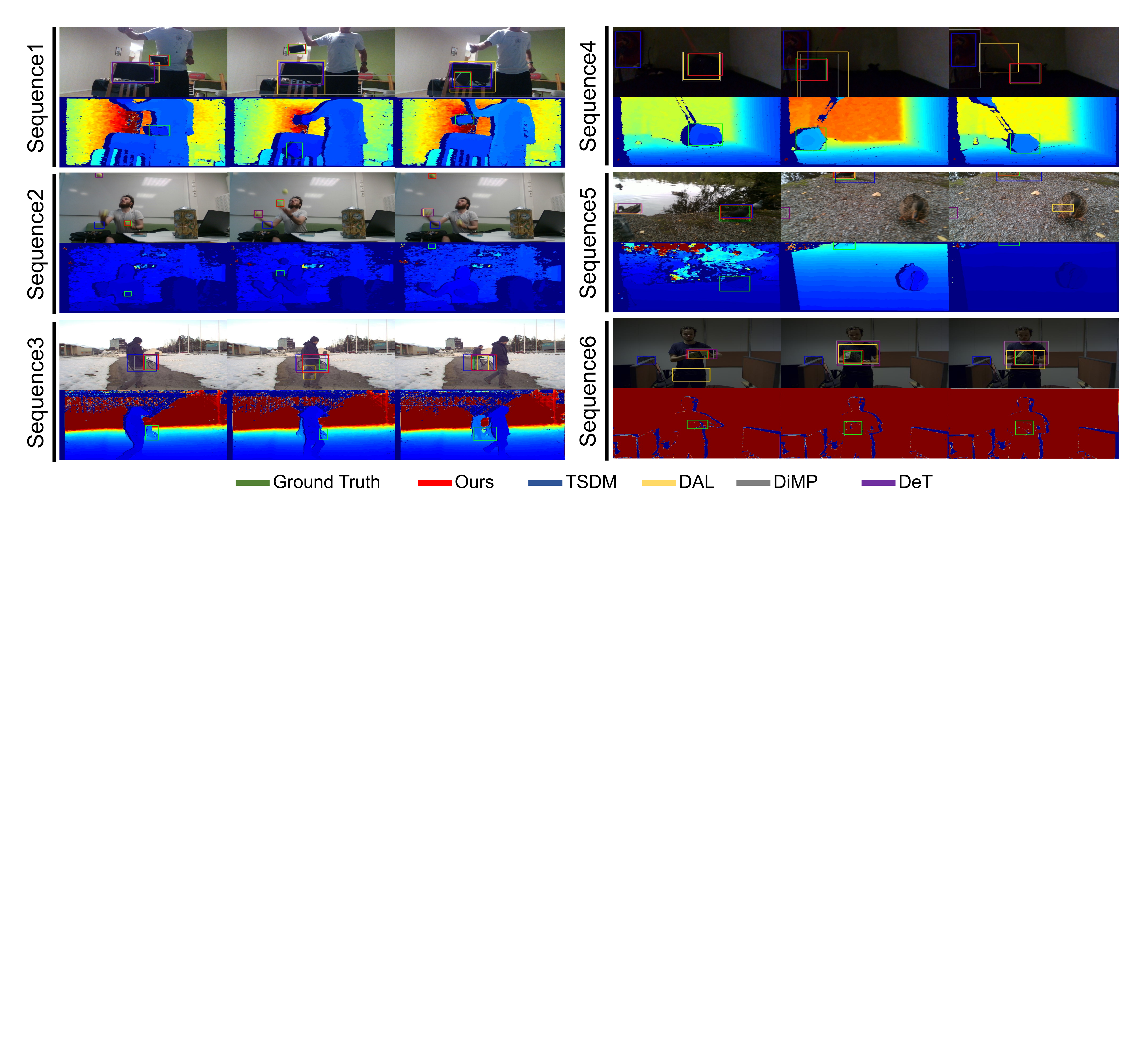}
\caption{Qualitative results in different challenging scenarios. Seq 1: background clutter; Seq 2: similar objects and fast motion; Seq 3: outdoor scenario and full occlusion; Seq 4: dark scenes and fast motion; Seq 5: outdoor scenario and camera motion; Seq 6: dark scene and target rotation.}
\label{results}
\end{figure*}

Figure \ref{results} shows several representative example results of representative trackers and the proposed tracker on various benchmark datasets.
Compared with other RGBD models, we can see that our DMTracker can achieve better tracking performance under multiple difficulties.
Sequence 1 shows a scene with background clutter. 
Our method can accurately distinguish the target object, while other trackers are confused by the clutters. 
In sequence 2, we show an example with multiple similar objects with fast motion, where it is challenging to accurately track the object. 
While our DMTracker can track the small-sized target ball robustly.
In sequence 3 and 5, we show two examples when there are background clutter or occlusion in outdoor scenarios. 
Our method locates the objects more accurately compared to other approaches. 
In sequence 4, under a very dark scene with fast motion of the object, our tracker can robustly track the target object.
Finally, It can be seen that some approaches fail to track the rotated object in a dark room, while our DMTracker can produce reliable results.
Overall, our tracker can handle various tracking challenges and produce promising results.



\begin{figure}[t]
  \centering
  \begin{minipage}{0.46\linewidth}
\centering
\includegraphics[width=\linewidth]{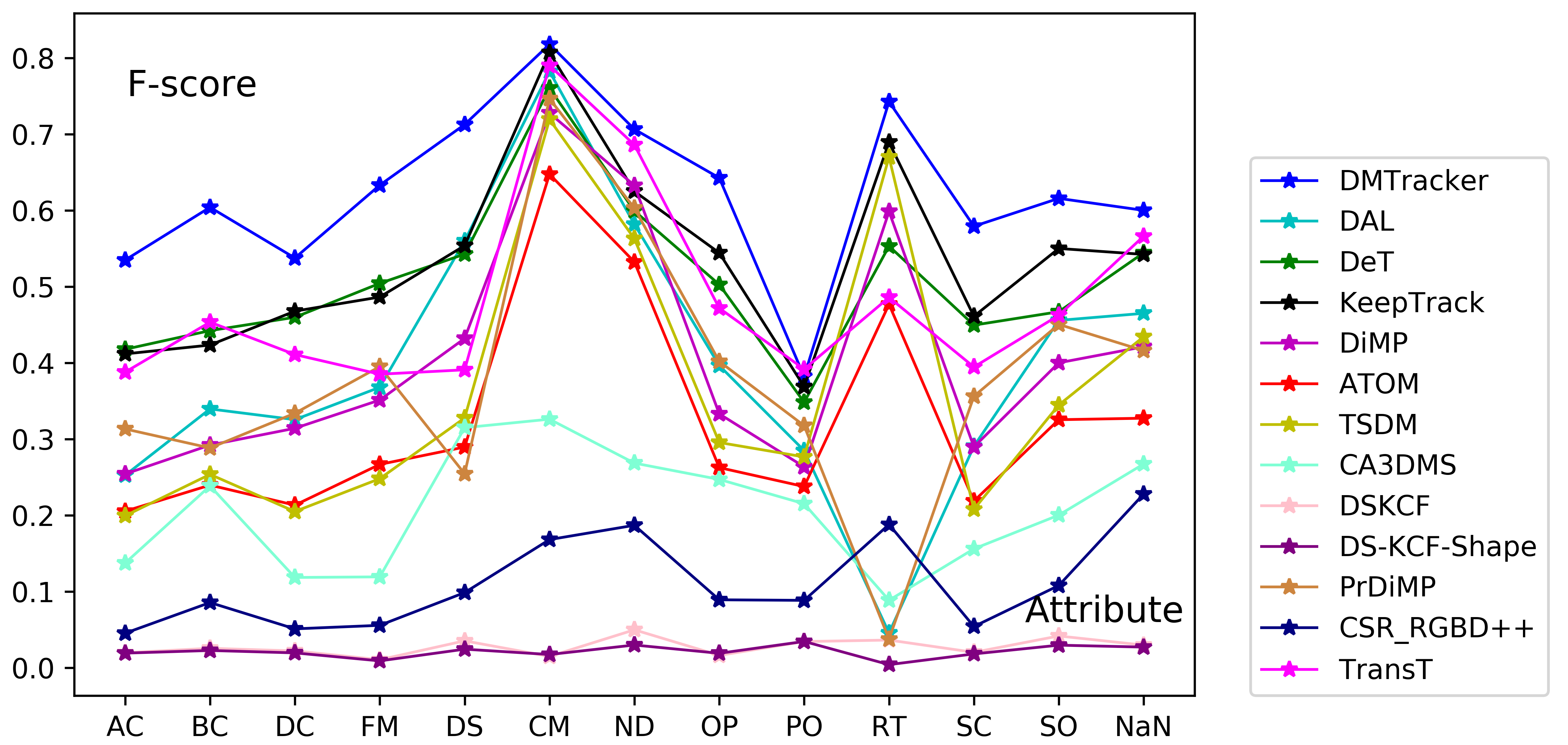}
\caption{Attribute-based F-score on DepthTrack \cite{det}.}
\label{depthtrack attribute}
  \end{minipage}
~  
  \begin{minipage}{0.46\linewidth}
\centering
\includegraphics[width=\linewidth]{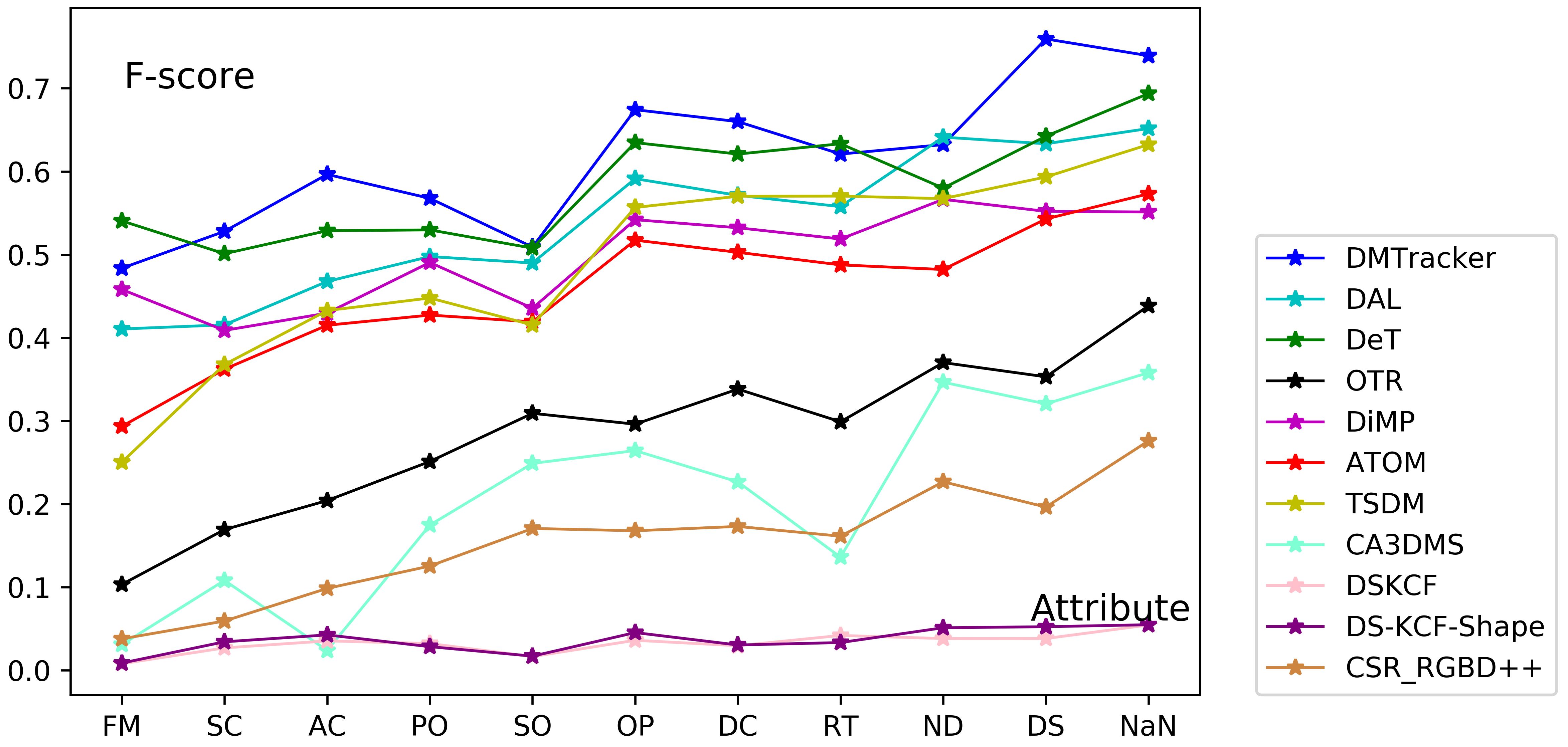}
\caption{Attribute-based F-score on CDTB \cite{cdtb}.}
\label{cdtb attribute}
  \end{minipage}
\end{figure}



\subsubsection{Attribute-based analysis.}
We report attribute-based evaluations of representative state-of-the-art algorithms on STC, DepthTrack, and CDTB for in-depth analysis. 
Success rates according to different attributes on STC are given in Table \ref{tbl_stc}.
We can see that our proposed outperforms other compared models on Color Distribution Variation (CDV), Depth Distribution Variation (DDV), Surrounding Color Clutter (SCC) , and Partial Occlusion (PO).
Attribute-specific F-scores for the tested trackers on DepthTrack and CDTB are shown in Figure \ref{depthtrack attribute} and Figure \ref{cdtb attribute}, respectively.
As shown, our DMTracker outperforms other trackers on 12 attributes on DepthTrack.
Our DMTracker outperforms other trackers on 7 attributes on CDTB.
It is obvious that DMTracker can handle the depth-related challenges better, \textit{e.g.}, background clutter (BC), dark scenes (DS), depth change (DC), and similar objects (SO).

\subsection{Ablation Study}

To verify the relative importance of different key components of our tracker, we conduct ablation studies 
on the most recent DepthTrack dataset \cite{det}.

\begin{table}
\caption{Ablation study on Cross-Modal Attention (CMA) layer numbers.}
\centering
\label{decoderlayer}
\begin{tabular}{l|c|c|c|c}
\hline
 Method & CMA Layer &\multicolumn{1}{c|}{Pr} & \multicolumn{1}{c|}{Re} & \multicolumn{1}{c}{F-score} \\
\hline
DMTracker & 1 & 0.585 & 0.571 &0.578\\
DMTracker & 2 & \textbf{0.619} & \textbf{0.597} & \textbf{0.608}\\
DMTracker & 3 & 0.570 & 0.552 &0.561\\
\hline
\end{tabular}
\end{table}




\subsubsection{Effects of CMA layers.}
To investigate the effects of different numbers of Cross-Modal Attention (CMA) layers, we compare tracking performance with different settings.
Table \ref{decoderlayer} shows the comparison results using different numbers of CMA layers, \textit{i.e.}, 1, 2, and 3. 
From the results, we can see that our model with 2 CMA layers obtains better performance. 
This can be explained that only one CMA layer leads to inadequate learning of the target knowledge and cross-modal correlation, while deep layers face the risks of overfitting.

\begin{table}[t]
\scriptsize
\begin{minipage}{0.42\linewidth}
\caption{Ablation study on key components in the network.}
\centering
\label{ab2}
\begin{tabular}{c c c |c|c|c}
\hline
 Base & CMIM & SPM &\multicolumn{1}{c|}{Pr} & \multicolumn{1}{c|}{Re} & \multicolumn{1}{c}{F-score} \\
\hline
\checkmark & &  & 0.517& 0.492  &0.504\\
\checkmark & \checkmark &   &0.522   &0.426  &0.469  \\
\checkmark & & \checkmark    &0.584 &0.563  &0.574\\
\checkmark & \checkmark & \checkmark  & 0.619 & 0.597 &0.608\\
\hline
\end{tabular}
\end{minipage}
~
\begin{minipage}{0.56\linewidth}
\caption{Ablation study on different SPM settings.}
\centering
\label{qkv}
\begin{tabular}{l|c|c|c|c|c}
\hline
 Method & $\alpha I^{0}+\beta F^{1}$ & $\alpha D^{0}+\beta F^{1}$  &Pr&Re & F-score \\
\hline
DMTracker &   & \checkmark &0.576 & 0.560 & 0.568\\
DMTracker & \checkmark  & & 0.619 & 0.597 &0.608\\
\hline
\end{tabular}
\end{minipage}
\end{table}

\begin{table}[t]
\scriptsize
\begin{minipage}{0.48\linewidth}
\caption{Overall performance on DepthTrack test set. The top 3 results are shown in {\color{red}red}, {\color{blue}blue}, and {\color{green}green}.}
\centering
\label{tbl_depthtrack}
\begin{tabular}{l|c|c|c|c}
\hline
 Method  & Type &\multicolumn{1}{c|}{Pr} & \multicolumn{1}{c|}{Re} & \multicolumn{1}{c}{F-score} \\
\hline
DS-KCF & RGBD &0.075&0.077&0.076 \\
DS-KCF-Shape& RGBD &0.070 &0.071 &0.071\\
CSR\_RGBD++  &RGBD & 0.113 &0.115 &0.114  \\
CA3DMS  &RGBD & 0.212 &0.216 &0.214\\
DAL  &RGBD &0.478 &0.390 &0.421 \\
TSDM  & RGBD & 0.393 &0.376 &0.384\\
DeT  & RGBD & \textcolor{blue}{0.560} & \textcolor{green}{0.506} &\textcolor{blue}{0.532} \\
DiMP50  & RGB & 0.387 &0.403 &0.395 \\
KeepTrack  & RGB &\textcolor{green}{0.504} &\textcolor{blue}{0.514} &\textcolor{green}{0.508} \\
TransT  & RGB &0.484 &0.494 &0.489 \\
\hline
\textbf{DMTracker} & RGBD &\textcolor{red}{0.619} &\textcolor{red}{0.597} &\textcolor{red}{0.608} \\
\hline
\end{tabular}
\end{minipage}
~
\begin{minipage}{0.5\linewidth}
\centering
\caption{Overall performance on CDTB dataset . The top 3 results are shown in {\color{red}red}, {\color{blue}blue}, and {\color{green}green}.}
\label{tbl_cdtb}
\begin{tabular}{l|c|c|c|c}
\hline
 Method  & Type &\multicolumn{1}{c|}{Pr} & \multicolumn{1}{c|}{Re} & \multicolumn{1}{c}{F-score} \\
\hline
DS-KCF &RGBD &0.036 &0.039 &0.038 \\
DS-KCF-Shape &RGBD &0.042 &0.044 &0.043\\
CA3DMS  &RGBD &0.271 &0.284 &0.259 \\ 
CSR\_RGBD++ &RGBD & 0.187 &0.201 &0.194 \\
OTR  &RGBD & 0.336 &0.364 &0.312\\
DAL &RGBD & \textcolor{blue}{0.661} &\textcolor{green}{0.565} &\textcolor{green}{0.592}\\
TSDM  & RGBD & 0.578 & 0.541 &0.559\\
DeT  & RGBD &\textcolor{green}{0.651}  &\textcolor{blue}{0.633} &\textcolor{blue}{0.642}\\
ATOM  & RGB & 0.541 &0.537 &0.539\\
DiMP50  & RGB & 0.546 &0.549 &0.547\\
\hline
\textbf{DMTracker} & RGBD &\textcolor{red}{0.662} &\textcolor{red}{0.658} &\textcolor{red}{0.660} \\
\hline
\end{tabular}
\end{minipage}
\end{table}

\subsubsection{Component-wise analysis.}
Since the proposed DMTracker is used to fuse cross-modal features and learn a unified representation, we ablate each key component for analysis.
The base model is constructed by removing the whole dual-fused modality-aware network.
The fused features are obtained by simply obtaining the element-wise addition of RGB and depth feature maps.
As shown in Table~\ref{ab2}, the tracking performance first degrades with only CMIM or SPM and then highly improves with adding the CMIM and SPM.
As we analyzed in Section \ref{visualization}, the CMIM will bring very informative and even redundant features which leads to the sub-optimal performance. With only SPM, the interaction of cross-modal information is lacking, so the model does not fully exploit the modality-aware correlation information.
Then, by adding the CMIM and SPM, the redundant information is effectively suppressed, shared and specificity information is enhanced and thus the tracking performance improves.

\subsubsection{Effects on modalities fusion order in SPM.}
In the specificity preserving module (SPM) of our DMTracker, We first add depth-specific features to RGBD features and then fuse them with weighted RGB-specific features. When given different orders of specific modalities, the performance of the tracker degrades. 
This can be explained by the fact that the RGB features are more informative than depth features, and this part of the information is dominant in the tracking process. Assume the spatial information is added in high weight at the end fusion. In that case, the proportion of visual information in the fusion feature is weakened. Mixing excess spatial information confuses the tracker.

\section{Conclusions}
In this paper, we proposed a novel DMTracker to learn dual-fused modality-aware representations for RGBD tracking.
On the one hand, the proposed cross-modal integration module can learn the shared information between RGB and depth channels through cross-modal attention.
On the other hand, the following specificity preserving module can adaptively fuse the shared features and RGB and depth features to get discriminative representations.
Extensive experiments validate the superior performance of the proposed DMTracker, as well as the effectiveness of each component.

\section*{Acknowledgements}
This work is supported by the National Natural Science Foundation of China under Grant No. 61972188 and 62122035.

\begin{table}[t]
\scriptsize
\begin{minipage}{0.48\linewidth}
\caption{Overall performance on DepthTrack test set. The top 3 results are shown in {\color{red}red}, {\color{blue}blue}, and {\color{green}green}.}
\centering
\label{tbl_depthtrack}
\begin{tabular}{l|c|c|c|c}
\hline
 Method  & Type &\multicolumn{1}{c|}{Pr} & \multicolumn{1}{c|}{Re} & \multicolumn{1}{c}{F-score} \\
\hline
DS-KCF & RGBD &0.075&0.077&0.076 \\
DS-KCF-Shape& RGBD &0.070 &0.071 &0.071\\
CSR\_RGBD++  &RGBD & 0.113 &0.115 &0.114  \\
CA3DMS  &RGBD & 0.212 &0.216 &0.214\\
DAL  &RGBD &0.478 &0.390 &0.421 \\
TSDM  & RGBD & 0.393 &0.376 &0.384\\
DeT  & RGBD & \textcolor{blue}{0.560} & \textcolor{green}{0.506} &\textcolor{blue}{0.532} \\
DiMP50  & RGB & 0.387 &0.403 &0.395 \\
KeepTrack  & RGB &\textcolor{green}{0.504} &\textcolor{blue}{0.514} &\textcolor{green}{0.508} \\
TransT  & RGB &0.484 &0.494 &0.489 \\
\hline
\textbf{DMTracker} & RGBD &\textcolor{red}{0.619} &\textcolor{red}{0.597} &\textcolor{red}{0.608} \\
\hline
\end{tabular}
\end{minipage}
~
\begin{minipage}{0.5\linewidth}
\centering
\caption{Overall performance on CDTB dataset . The top 3 results are shown in {\color{red}red}, {\color{blue}blue}, and {\color{green}green}.}
\label{tbl_cdtb}
\begin{tabular}{l|c|c|c|c}
\hline
 Method  & Type &\multicolumn{1}{c|}{Pr} & \multicolumn{1}{c|}{Re} & \multicolumn{1}{c}{F-score} \\
\hline
DS-KCF &RGBD &0.036 &0.039 &0.038 \\
DS-KCF-Shape &RGBD &0.042 &0.044 &0.043\\
CA3DMS  &RGBD &0.271 &0.284 &0.259 \\ 
CSR\_RGBD++ &RGBD & 0.187 &0.201 &0.194 \\
OTR  &RGBD & 0.336 &0.364 &0.312\\
DAL &RGBD & \textcolor{blue}{0.661} &\textcolor{green}{0.565} &\textcolor{green}{0.592}\\
TSDM  & RGBD & 0.578 & 0.541 &0.559\\
DeT  & RGBD &\textcolor{green}{0.651}  &\textcolor{blue}{0.633} &\textcolor{blue}{0.642}\\
ATOM  & RGB & 0.541 &0.537 &0.539\\
DiMP50  & RGB & 0.546 &0.549 &0.547\\
\hline
\textbf{DMTracker} & RGBD &\textcolor{red}{0.662} &\textcolor{red}{0.658} &\textcolor{red}{0.660} \\
\hline
\end{tabular}
\end{minipage}
\end{table}

\begin{table*}
\scriptsize
\centering
\caption{Overall performance and attribute-based results on STC . The top 3 results are shown in {\color{red}red}, {\color{blue}blue}, and {\color{green}green}.}\label{tbl_stc}
\begin{tabular}{l |c |c | cccccccccc}
\hline
&  & & \multicolumn{10}{c}{Attributes} \\
Method & Type &Success &\multicolumn{1}{c}{IV} & \multicolumn{1}{c}{DV} & \multicolumn{1}{c}{SV} & \multicolumn{1}{c}{CDV} & \multicolumn{1}{c}{DDV} & \multicolumn{1}{c}{SDC} & \multicolumn{1}{c}{SCC} & \multicolumn{1}{c}{BCC} & \multicolumn{1}{c}{BSC} & \multicolumn{1}{c}{PO} \\
\hline
OAPF & RGBD &0.26&0.15&0.21&0.15&0.15&0.18&0.24&0.29&0.18&0.23&0.28\\
DS-KCF & RGBD &0.34&0.26&0.34&0.16&0.07&0.20&0.38&0.39&0.23&0.25&0.29\\
PT  & RGBD &0.35&0.20&0.32&0.13&0.02&0.17&0.32&0.39&0.27&0.27&0.30\\
DS-KCF-Shape  & RGBD &0.39&0.29&0.38&0.21&0.04&0.25&0.38&0.47&0.27&0.31&0.37\\
STC  & RGBD &0.40&0.28&0.36&0.24&0.24&0.36&0.38&0.45&0.32&0.34&0.37\\
CA3DMS  & RGBD &0.43&0.25&0.39&0.29&0.17&0.33&0.41&0.48&0.35&0.39&0.44\\
CSR\_RGBD++ & RGBD &0.45&0.35&0.43&0.30&0.14&0.39&0.40&0.43&0.38&0.40&0.46\\
OTR  & RGBD &0.49&0.39&0.48&0.31&0.19&0.45&0.44&0.46&0.42&0.42&0.50\\
DAL  & RGBD &\textcolor{blue}{0.62} &\textcolor{blue}{0.52}  &0.60 &\textcolor{red}{0.51} &	\textcolor{blue}{0.62} &	0.46 &	\textcolor{blue}{0.63} &	\textcolor{green}{0.63} &	\textcolor{red}{0.55} &	\textcolor{red}{0.58} &	0.57 \\
TSDM  & RGBD &0.57 &\textcolor{red}{0.53}  &0.42 &0.43 &	0.54 &	0.37 &	0.56 &	0.55 &	0.41 &	0.45 &	0.53 \\
DeT   & RGBD &0.60 &0.39  &0.56 &\textcolor{blue}{0.48} &	\textcolor{green}{0.58} &	0.40 &	0.59 &	0.62 &	0.48 &	0.46 &	0.54 \\ 
DiMP & RGB &0.60 &0.49 &\textcolor{green}{0.61} &\textcolor{green}{0.47} &0.56 &\textcolor{green}{0.57} &0.60 &\textcolor{blue}{0.64} &0.51 &\textcolor{blue}{0.54} &0.57 \\
PrDiMP  & RGB &0.60 &0.47 &\textcolor{blue}{0.62} &	\textcolor{green}{0.47} &	0.51 &	\textcolor{blue}{0.58} &	\textcolor{red}{0.64} &	0.62 &	\textcolor{blue}{0.53} &	\textcolor{blue}{0.54} & \textcolor{blue}{0.58} \\
KeepTrack & RGB & \textcolor{blue}{0.62} & 0.51 &\textcolor{red}{0.63} &	\textcolor{green}{0.47} &	0.57 &	0.56 &	\textcolor{green}{0.61} &	0.62 &	\textcolor{blue}{0.53} &	\textcolor{green}{0.52} & \textcolor{blue}{0.58}  \\
\hline
\textbf{DMTracker} & RGBD & \jinyu{0.63} & \textcolor{blue}{0.52} &0.59  &0.46 &\jinyu{0.67} &\jinyu{0.63} &	\textcolor{green}{0.61} &	\jinyu{0.65} &	\textcolor{green}{0.52} &	\textcolor{green}{0.52} &\jinyu{0.60} \\
\hline
\end{tabular}
\end{table*}



\clearpage
%
%
\bibliographystyle{splncs04}
\bibliography{egbib}
\end{document}